\documentclass[letterpaper, 10 pt, conference]{ieeeconf}  

\IEEEoverridecommandlockouts                              

                                    \usepackage{graphicx}
\usepackage{placeins}
\usepackage{float}
\usepackage{amsmath}
\usepackage{subfigure}
\usepackage{wrapfig}
\usepackage{multirow}
\usepackage{lipsum,blindtext}
\usepackage{adjustbox}
\newcommand{\norm}[1]{\left\lVert#1\right\rVert}

\DeclareMathOperator*{\argmin}{arg\,min}
\usepackage{soul,color}

\usepackage{amsfonts}

\title{Social-STAGE: Spatio-Temporal Multi-Modal \\Future Trajectory Forecast}

\author{Srikanth Malla \and Chiho Choi \and Behzad Dariush\thanks{The authors are with Honda Research Institute USA, 70 Rio Robles, San Jose, CA 95134, USA.
{\tt\small \{smalla,cchoi, bdariush\}@honda-ri.com}}
}

\begin{document}

\maketitle
\thispagestyle{empty}
\pagestyle{empty}


\begin{abstract}
This paper considers the problem of multi-modal future trajectory forecast with ranking. 
Here, multi-modality and ranking refer to the multiple plausible path predictions and the confidence in those predictions, respectively.  
We propose Social-STAGE, \textbf{Social} interaction-aware \textbf{S}patio-\textbf{T}emporal multi-\textbf{A}ttention \textbf{G}raph convolution network with novel \textbf{E}valuation for multi-modality. 
Our main contributions include analysis and formulation of multi-modality with ranking using interaction and multi-attention, and introduction of new metrics to evaluate the diversity and associated confidence of multi-modal predictions.
We evaluate our approach on existing public datasets ETH and UCY and show that the proposed algorithm outperforms the state of the arts on these datasets.
\end{abstract}



\section{Introduction}


Forecasting the trajectory of agents in dynamic scenes is an important research problem with a range of applications such as autonomous navigation, surveillance, and human-robot interaction. The challenge in addressing this problem lies in modeling the variability and uncertainty of human behavior (\textit{i.e.}, multi-modality) and the associated social and cultural norms, particularly in a highly unstructured environment that involves complex interactions between agents (\textit{i.e.}, social interaction). 

The importance of interaction modeling and multi-modality has been highlighted by studies on interaction-aware multi-modal trajectory prediction~\cite{lee2017desire,gupta2018social,vemula2018social,cui2019multimodal,malla2020titan}, where they extract the interaction-encoded feature representations from past observation and train a deep neural network to generate a certain distribution of future trajectories. Although different types of distributions\footnote{Examples of the learned distribution are as follows: bi-variate Gaussian distribution~\cite{alahi2016social,vemula2018social, mohamed2020social, malla2020titan}, mixture models~\cite{curro2018deriving,messaoud2018structural,cui2019multimodal,makansi2019overcoming}, multi-variate normal distribution~\cite{lee2017desire,choi2019drogon,choi2020shared}, or true data distribution~\cite{gupta2018social,kosaraju2019social}} are learned, the predicted outputs are likely being a single mode with high variances~\cite{makansi2019overcoming} or mode-collapse providing limited mode variety~\cite{mohamed2020social}. In addition, most existing multi-modal methods do not consider the probability of each mode and their ranking, which places strict limits on their practical use in safety critical applications such as autonomous navigation for robotics or driving assistance systems. For example, consider a case where a prediction algorithm generates multiple future trajectories of an agent for robot navigation. Most existing multi-modal methods cannot specify the best prediction since they do not measure a probability of individual modes nor ranking trajectories. This limitation makes it difficult for the system to make a proper decision with the given multi-modal prediction results.

\begin{figure}[t]
\centering
\includegraphics[scale=.45]{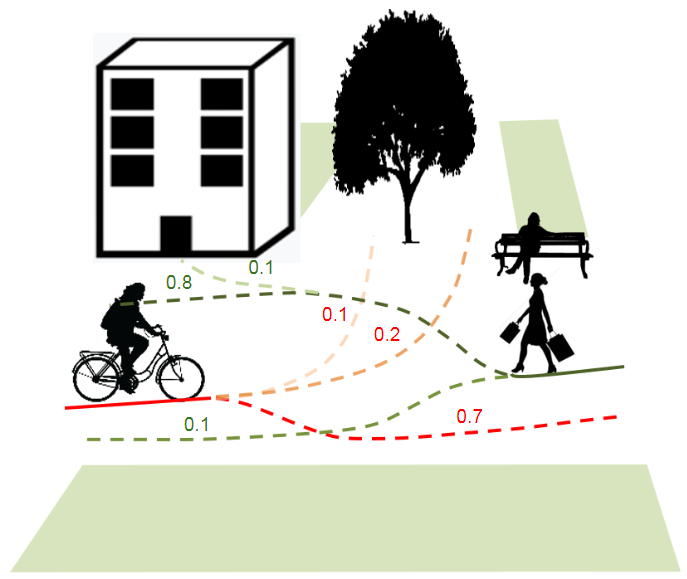}
\caption{
Two agents approaching a junction from opposite directions. Past motion history is shown in solid line, and multiple plausible future trajectories are shown in dashed line. Future paths could evolve with varying probability/ranking as depicted with different opacity of the dashed line. The evolution of each mode can be characterized by voluntary movements toward several possible directions or destinations, or from interactive or reactive behaviors that evolve from negotiation between the two agents which may also include social norms or rules.
}
\label{fig:myfig}
\end{figure}
In this work, we propose Social-STAGE, \textbf{Social} interaction-aware \textbf{S}patio-\textbf{T}emporal multi-\textbf{A}ttention \textbf{G}raph convolution network with novel \textbf{E}valuation for multi-modality.  
Our model predicts plausible future trajectories with a diversity, considering spatio-temporal interactions of agents. An overview of our approach is illustrated in Fig. 2. As a first step, we model social interactions between humans using a spatio-temporal graph convolutional network~\cite{yan2018spatial}. Given the observation of agents' motion history, we simultaneously explore the spatial influences of individual entities and their temporal changes, creating spatio-temporal interactions. Then, we highlight more important interactions in space and in time in the multi-attention module that follows. The resulting interaction features are decoded as a set of plausible deterministic trajectories with corresponding probabilities. 

Average distance error (ADE) and final distance error (FDE) metrics are standard evaluation metrics used in trajectory prediction. Although they clearly quantify the distance error between the predicted locations and those of the ground-truth, we believe that the prediction capability of multi-modal approaches cannot be properly evaluated. The issues observed from these metrics are as follows: (i) Oracle error has been used over a set of prediction samples, only considering the best prediction result. Since the remaining prediction results are not at all evaluated, this process overlooks the diversity of multi-modal trajectories. (ii) The standard ADE and FDE metrics do not consider any type of prediction confidence such as a probability or rank of each trajectory. In practice, however, such a confidence factor should be provided and evaluated in applications where the ground-truth is not available during inference. Considering these limitations, we additionally introduce new error metrics based on ADE and FDE to properly evaluate the diversity of multi-modal predictions with their associated confidence.

\section{Related Works}
\subsection{Deep Learning on Trajectory Forecast}
Deep neural networks have been used with great success in modeling interactions between humans in a data-driven manner. A recurrent neural network (RNN) has been widely used to encode the history of pedestrians' motion~\cite{alahi2016social,lee2017desire,gupta2018social,xu2018encoding,zhang2019sr}.  RNNs use a pooling mechanism to extract interaction information by aggregating the encoded motion features of different individuals. However, such hand-crafted or arbitrary aggregation methods may hinder the acquisition of important representations for future trajectory generation. To address this limitation, a graph neural network has been introduced in this domain. The inherent node-edge topology of graphs helps to model more intuitive and natural interactions between humans \cite{vemula2018social,ma2019trafficpredict,kosaraju2019social,mohamed2020social}. Rather than pooling the information, these approaches find interactions between agents, corresponding to updating edges. However, they are prone to error accumulation over the time horizon because of their structural dependency on recurrent units~\cite{vemula2018social,ma2019trafficpredict}, limited in modeling interactions in space and in time~\cite{kosaraju2019social}, or do not properly capture the importance of interactions in a spatio-temporal space~\cite{mohamed2020social}. In contrast, our aim in this work is to propose an approach that can overcome such limitations using graph neural networks~\cite{kipf2016semi} with multi-attention in space and in time.

\subsection{Multi-Modal Trajectory Forecast}
The importance of multi-modality in trajectory prediction has become more widely recognized in recent years. Considering that multiple plausible trajectories may
exist with the given information, some works~\cite{alahi2016social,vemula2018social, mohamed2020social, malla2020titan} generate a bi-variate Gaussian distribution and sequentially sample the pedestrian's future locations. Other works address the multi-modality of the future prediction using mixture models~\cite{hjorth1999regularisation,graves2013generating,rupprecht2017learning,curro2018deriving,messaoud2018structural,cui2019multimodal}. However, their predictions tend to be a single mode with high variances~\cite{makansi2019overcoming}. There are also methods that extend deep generative models to learn the distribution of future trajectories over data. These methods use either variational autoencoders (VAEs)~\cite{lee2017desire,choi2019drogon,choi2020shared} or generative adversarial networks (GANs)~\cite{gupta2018social,sadeghian2019sophie}. Although they are expected to generate multi-modal human behavior, such random sampling-based strategies are prone to experiencing problems with mode collapse~\cite{mohamed2020social} or posterior collapse~\cite{choi2020shared}. More importantly, those works do not measure the probability of individual modes, and therefore, their multi-modal predictions have practical limitations in many real applications. 
\cite{chai2020multipath} recently addresses more practical functionality of multi-modality by providing the probability of each trajectory output. However, a predefined trajectory set is used to represent the different modes of future motion, which is difficult to generalize across different environment types. In this work, our model not only generates interaction-specific multi-modal future trajectories with diversity, it also outputs probabilities associated with individual modes.

\begin{figure*}[t!]
    \centering
    \includegraphics[scale=0.45]{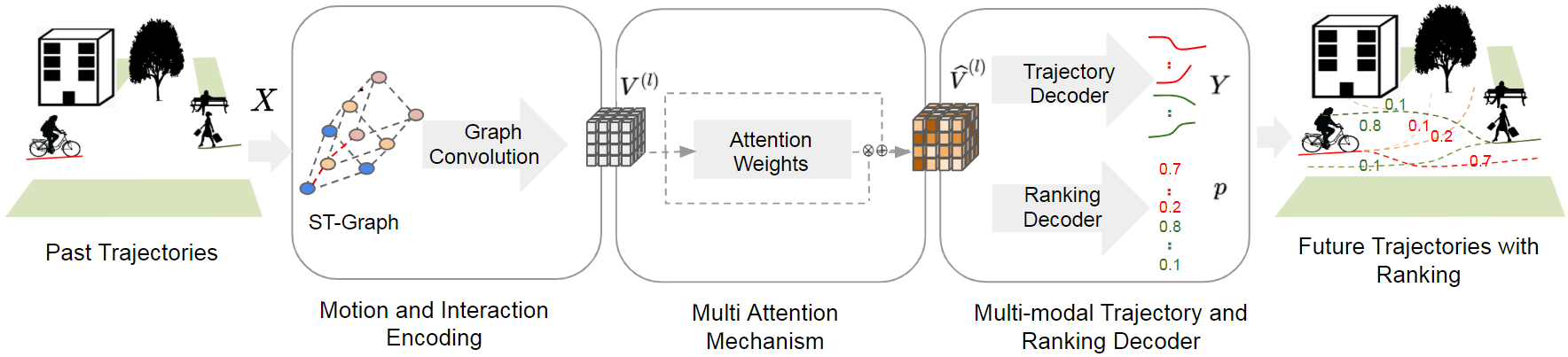}
    \caption{Social-STAGE Framework: Given input scenario with agents' (shown in different colors) past trajectories (shown in dotted line), we perform ST-Graph Convolutions and multi-attention mechanism for encoding meaningful interactions. The encoded features are decoded into multi-modal trajectories (for all agents) and corresponding probabilities to rank each mode. The opacity of each future trajectory prediction is changed to show the corresponding probability in the final output.}
    \label{fig:framework}
    \vspace{-0.5cm}
\end{figure*}

\subsection{Evaluation Metrics for Trajectory Forecast}
Evaluation of future trajectory has been conducted by computing Euclidean ($L_2$) distance between the predicted and ground-truth locations at every future time step. The error is reported by measuring the average distance error (ADE) during a specified time interval,  and the final distance error (FDE) at a specified time. Since these metrics are intuitive and directly quantify the performance of prediction algorithms, they are generally used in evaluation of performance in previous approaches~\cite{alahi2016social,vemula2018social}. For multi-modal evaluation, \cite{lee2017desire} simply extends the same metrics in an oracle manner by evaluating one with the lowest error.   Their method computes the performance upper bound instead of evaluating the diversity or plausibility of multiple modes. \cite{ivanovic2019trajectron,thiede2019analyzing} introduce the kernel density estimate-based negative log-likelihood (KDE\_NLL) metric that computes the log-likelihood of the ground-truth against the sampled trajectory distribution.  
The KDE function often underestimates the output distribution, approximating as a Gaussian kernel. When multiple trajectory modes are predicted (e.g., different trajectories at an intersection), it does not properly capture the multi-modality because of the kernel approximation. In contrast, the proposed metrics avoid this potential problem without the approximation and directly evaluate the diversity of multi-modal trajectories with associated probabilities.

\section{Methodology}
\subsection{Problem Formulation}
Given a scenario with $K$ agents and their motion information, we use the first $T_{in}$ time steps of trajectories $X=\{X^1,X^2,...,X^K\}$ to predict the next $T_{out}$ time steps of trajectories $Y=\{Y^1,Y^2,...,Y^k\}$. We denote $X^k=\{x_t^k|\forall t\in\{1,...,T_{in}\}\}$ and $Y^k=\{y_{t}^k|\forall t\in\{1,...,T_{out}\}\}$ for each agent $k\in \{1,...,K\}$. $x_t^k$ and $y_t^k$ can be further represented by 2D positions of agent $k$ at time $t$. Our framework outputs $M$ modes of future trajectories $\hat{Y}^k_{m}=\{\hat{y}_{(m)1}^k,\hat{y}_{(m)2}^k,..\hat{y}_{(m)T_{out}}^k\}$ with their probabilities $p^k_m$ for all $K$ agents in the scene, where $m\in\{1,...,M\}$, such that $\sum_{m} p^k_m=1$.

\subsection{Spatio-Temporal Multi-Attention Graph Convolutions}
We create a graph representation $G_t=(V_t,E_t)$ at each time step $t\in \{1,...,T_{in}\}$, where $V_t=\{v_t^k|\forall k\in\{1,...,K\}\}$ is a set of nodes and $E_t=\{e_t^{ij}|\forall i,j\in\{1,...,K\}\}$ is a set of edges.  $v^k_t= x^k_t-x^k_{t-1}$ is a node attribute represented as a relative motion for agent $k$, which also encodes the heading at each time-step. Also, $e_t^{ij}$ is an edge attribute where $e_t^{ij}$ indicates the connectivity between two node attributes. 

We follow the procedure as in~\cite{kipf2016semi,yan2018spatial} for the implementation of graph convolutions. We define the adjacency matrix $A_t$ where its element $a_t^{ij}$ indicates the importance weight of the edge between two node attributes $v_t^i$ and $v_t^j$. The weight is represented using a kernel function similar to~\cite{mohamed2020social}, $a^{ij}_t=1 / \norm{v^i_t-v^j_t}_2$ if $i\neq j$, otherwise $a^{ii}_t=0$. To ease training, we adopt a renormalization trick. Therefore, the adjacency matrix is symmetrically normalized as 
\begin{equation}
A_t = \lambda_t^{-1/2}\hat{A_t}\lambda_t^{-1/2}    
\end{equation}
where $\lambda_t$ is a diagonal node degree matrix of $A_t$ and $\hat{A_t}=A_t+I$ adds self-connections to $A_t$ with an identity matrix $I$. For spatio-temporal extension, we stack the adjacency matrices $A = \{A_t|\forall t\in \{1,...,T_{in}\}\}$ and nodes $V = \{V_t|\forall t\in \{1,...,T_{in}\}\}$ for all observation time steps. In addition, $\hat{A}$ is stack of $\hat{A_t}$ and $\lambda$ is stack of $\lambda_t$. Therefore, the node attributes $V^{(l)}$ of layer $l$ are updated using the adjacency matrix as follows:
\begin{equation}
    f(V^{(l)}, A) = \sigma(\lambda^{-1/2}\hat{A}\lambda^{-1/2}V^{(l)}\textbf{W}^{(l)}),
\label{eq:st-graph-conv}
\end{equation}
where $\sigma$ is an activation function and \textbf{W}$^{(l)}$ is the trainable weights of layer $l$. 





\begin{figure}[h]
    \centering
    \includegraphics[scale=0.35]{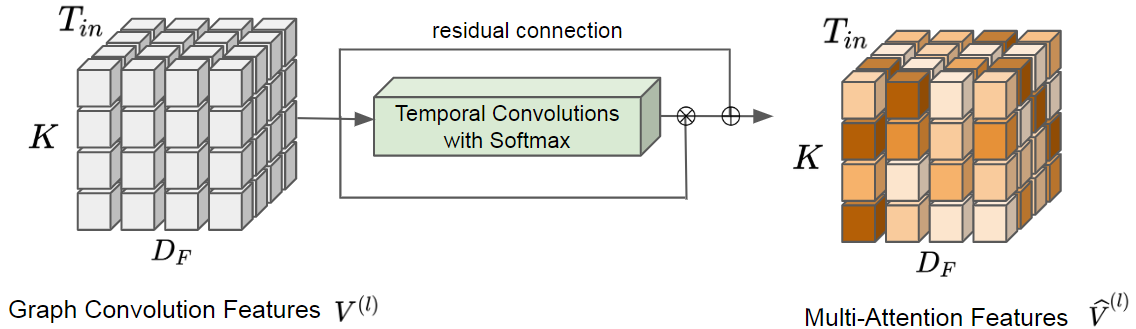}
    \caption{Multi Attention: $\oplus$ and $\otimes$ are element-wise sum and multiplication respectively}
    \label{fig:multi-attn}
    \vspace{-0.5cm}
\end{figure}

Therefore, the resulting node attributes are formulated as graph convolutional features for spatio-temporal interactions. We further use temporal convolutions and the softmax operation to obtain the attention weights. Such weights correspond to the relative importance of interactions, describing the time when the interaction should be captured and the agent who should be identified. We use the term multi-attention as individual agents can be simultaneously attentive along the spatial and temporal dimension. Note that this process is different with prior works~\cite{vemula2018social,ma2019trafficpredict}, where a single attentive weight is locally generated in space with a recurrent fashion. As a result, we can capture social interactions activated across different spatial and temporal locations. The multi-attention weights are updated by the residual connection similar to \cite{vaswani2017attention}. This leaves us with the following operation:
    \begin{equation}
        \hat{V}^{(l)} = (\phi(V^{(l)}) \otimes V^{(l)}) \oplus V^{(l)},
        \label{eq:attn}
    \end{equation}
where $\phi$ denotes a convolution operation to compute attention weights, $\otimes$ denotes element-wise multiplication, and $\oplus$ denotes element-wise sum operation. The updated features $\hat{V}^{(l)}$ are multi-attention features with attentive weights of spatio-temporal interactions. This process is visualized in Figure~\ref{fig:multi-attn}, and the detailed network architecture is presented in the Section~\ref{sec:implementation_model}.
\subsection{Ranking and Decoding}
\begin{figure}[t]
    \centering
    \includegraphics[scale=0.31]{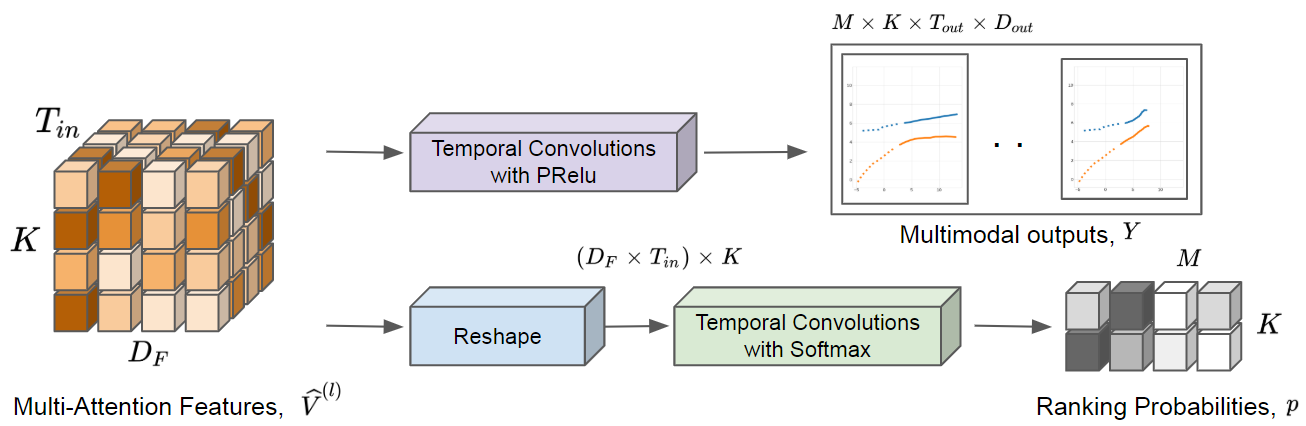}
    \caption{Decoding multi modal trajectories and probabilities for each mode and agent}
    \label{fig:decoder}
    \vspace{-0.5cm}
\end{figure}
To address the practical usage of multi-modality, we introduce probability measures for each output trajectory mode to rank the confidence in our predictions. This step is displayed in Figure~\ref{fig:decoder}. The first stream is constructed with temporal convolutions with a PRelu operation in the decoder to generate $M$ future trajectories of $K$ agents for $T_{out}$ time steps. Therefore, each output trajectory matches to one of the modes. Since in our implementation the dimension of multi-attention features $D_{F}$ is same as the dimension of the output trajectory $D_{out}$, no additional operations are necessary to match the dimensions. The second stream generates probabilities corresponding to individual output modes. We reshape the tensor by combining the first two dimensions ($D_F$, $T_{in}$) and then perform temporal convolutions on the combined dimension with a softmax operation across the output modes 
dimension $M$ as shown in Figure~\ref{fig:decoder}.

The total loss is shown in Eqn.~\ref{eq:total_loss}. $L_{ce}$ is the cross entropy loss from ranking the modes by their predicted probabilities. $L^{min}_{reg}$ is the reconstruction loss for predicting the trajectories.
\begin{equation}
L = L_{ce} + L^{min}_{reg}
\label{eq:total_loss}
\end{equation}
During training, we predict all $M$ output modes. However, we only have one ground truth trajectory per agent, and thus use $L^{min}_{reg}$.  Similar to the variety loss in \cite{gupta2018social} for penalizing multi-modal outputs, we penalize the minimum error mode $L_{reg}^{min}$ out of all output modes to preserve multi-modality for all agents as shown in Eqn.~\ref{eq:min_error}. $m_{min}$ is the minimum error mode, $Y^k$ is the ground-truth future trajectory, and $\hat{Y}^{k}_{m}$ is a prediction mode ($m$) of agent $k$.
\begin{equation}
\begin{split}
L^{min}_{reg}&=\sum_kL_{reg}(Y^{k}_{m_{min}})\\
m_{min} &= \argmin_m L_{reg}(Y^{k}_{m})\\
L_{reg}(Y^{k}_{m}) &= \norm{Y^k -\hat{Y}^{k}_{m}}_2\\
\end{split}
\label{eq:min_error}
\end{equation}
Since there is no ground truth for probabilities associated with each mode, we penalize each mode in an unsupervised manner as shown in Eqn.~\ref{eq:prob_loss}, where $p^{k,m}_{gt}$ is the ground-truth probability and $p^{k,m}_{pred}$ is the prediction probability for mode $m$ of agent $k$. $p^{k,m}_{gt}$ is generated using the minimum error criteria. We find the ground-truth probability of each prediction mode based on its proximity to the ground truth trajectory. Finally, $ce$ is cross entropy  for classification.  
\begin{equation}
\begin{split}
L_{ce} &= \sum_k \sum_m ce(p^{k,m}_{gt}, p^{k,m}_{pred})\\
p^{k,m}_{gt} &=
\begin{cases}
1 &, m=m_{min}\\
0 &, \textit{otherwise}
\end{cases}
\end{split}
\label{eq:prob_loss}
\end{equation}

\subsection{Evaluation Metrics}
Evaluation of future trajectory is conducted by computing the Euclidean (L2) distance between the predicted and ground-truth locations. The standard error metrics are Average Displacement Error (\textit{ADE}), Final Displacement Error (\textit{FDE}) for single-modal prediction, and minimum Average Displacement Error (\textit{ADE}$_{min}$), minimum Final Displacement Error (\textit{FDE}$_{min}$) for multi-modal prediction with $M$ output modes:
\begin{equation}
\begin{split}
    ADE &= \frac{1}{T_{out}}\sum_t \norm{x_t-\hat{x}_t}_2\\
    FDE &= \norm{x_T-\hat{x}_T}_2\\
    ADE_{min} &= \min_m ADE(m)~~~~\forall m\in\{1,...,M\}\\
    FDE_{min} &= \min_m FDE(m)~~~~\forall m\in\{1,...,M\}\\
\end{split}
\label{eq:metrics_old}
\end{equation}
Although these metrics quantify the performance of prediction algorithms, the oracle use of such metrics to evaluate multi-modality has the following inherent limitations: (i) considering only one trajectory with the lowest error, these metrics compute the performance upper bound and simply overlook the diversity of multiple modes; and (ii) these metrics do not take a prediction confidence into account if the associated probability were available.  To address these limitations, we introduce new error metrics that can evaluate the diversity of multi-modal predictions with their associated confidence.

For any selected mode $\hat{m}$ (using some criteria like mean or maximum probability), we find the error contributed by other modes $\mathcal{M} = \mathrm{E}(e_i)- \hat{p}*\hat{e}$, excluding error $\hat{e}$ contributed by probability $\hat{p}$ of mode ($\hat{m}$) from expectation of all modes errors $\mathrm{E}(e_i)$.
Based on this, we introduce two new metrics that do not make any assumption on the method’s output distribution: (i) The $\mathcal{M}_1$ metric is straightforward to compute the diversity of trajectories with respect to the ground-truth as we subtract the error contributed by the best mode ($\hat{e}$) from the expectation of errors of all modes by assigning equal probability ($1/M$) to all modes including the best mode ($\hat{p}=1/M$). Therefore, we see how diverse the predicted samples are on average for the given test dataset. (ii) The $\mathcal{M}_2$ metric is designed to evaluate the error contribution of other modes with respect to their confidence. We subtract weighted error of the best mode ($\hat{e} = e_{p_{max}}$, maximum probability mode error with $\hat{p}=p_{max}$) from the weighted expectation of errors of all modes using the probabilities predicted. Therefore, we measure how distributed the predicted samples are with respect to their probability. With lower values, the prediction model is likely to be certain about its output with high accuracy. 

\begin{equation}
\begin{split}
    \mathcal{M}_1 &= \frac{1}{M}((\sum_ie_i)- \hat{e})\\ 
    \mathcal{M}_2 &= (\sum_ip_i*e_i)- p_{max}*e_{p_{max}}  
\end{split}
\label{eq:metrics_new}
\end{equation}

\begin{table*}[!t]
\centering\small
\begin{tabular}{ c|c|c|c|c|c|c}
\hline
&ETH&Hotel&Univ&Zara1&Zara2&Avg\\\hline
LSTM&1.09 / 2.41& 0.86 / 1.91&0.61 / 1.31&0.41 / 0.88&0.52 / 1.11&0.70 / 1.52\\
S-LSTM~\cite{alahi2016social}&1.09 / 2.35&0.79 / 1.76 &0.67 / 1.40&0.47 / 1.00&0.56 / 1.17&0.72 / 1.54\\
S-Attn~\cite{vemula2018social}&1.39 / 2.39&2.51 / 2.91&1.25 / 2.54&1.01 / 2.17&0.88 / 1.75&1.41 / 2.35\\
CIDNN~\cite{xu2018encoding}&1.25 / 2.32&1.31 / 2.36&0.90 / 1.86&0.50 / 1.04&0.51 / 1.07&0.89 / 1.73\\
SGAN~\cite{gupta2018social}&1.13 / 2.21&1.01 / 2.18&0.60 / 1.28&0.42 / 0.91&0.52 / 1.11&0.74 / 1.54\\
S-STGCNN~\cite{mohamed2020social}&0.92 / 1.81& 0.76 / 1.49& 0.63 / 1.26&0.52 / 1.06&0.44 / 0.90&0.65 / 1.30\\
STGAT~\cite{huang2019stgat}&{0.88} / \textbf{1.66} &0.56 / 1.15&\textbf{0.52} / \textbf{1.13}&0.41 / 0.91&{0.31} / \textbf{0.68}&0.54 / {1.11}\\\hline
{S-STAGE (ours-single)} & \textbf{0.88} / 1.87 & \textbf{0.44} / \textbf{0.95} &0.53 / 1.15& \textbf{0.39} / \textbf{0.87} & \textbf{0.31} / 0.69 & \textbf{0.51} / \textbf{1.11}\\
\hline\hline
\multicolumn{7}{c}{Maximum Probability Mode Error $\downarrow$}\\\hline
{S-STAGE (ours-pmax)} &\multirow{2}{*}{\textbf{0.75 / 1.55}}&\multirow{2}{*}{\textbf{0.43 / 0.86}}&\multirow{2}{*}{0.53 / 1.15}&\multirow{2}{*}{\textbf{0.39 / 0.87}}&\multirow{2}{*}{\textbf{0.31} / 0.69}&\multirow{2}{*}{\textbf{0.48 / 1.02}}\\
$m_{p_{max}}$:[5,3,1,1,1]&&&&&\\
\hline
\end{tabular}
\caption{Single-modal ADE / FDE for $T_{pred}$ = 12 timesteps are reported in meters.}
\label{tbl:results-single}
\end{table*}

\begin{table*}[!t]
\centering\small
\begin{tabular}{ c|c|c|c|c|c|c}

\hline
&ETH&Hotel&Univ&Zara1&Zara2&Avg\\\hline

SGAN-20~\cite{gupta2018social} &0.81 / 1.52&0.72 / 1.61&0.60 / 1.26&0.34 / 0.69&0.42 / 0.84&0.58 / 1.18\\
Sophie-30~\cite{sadeghian2019sophie} &0.70 / 1.43&0.76 / 1.67&0.54 / 1.24&0.30 / 0.63&0.38 / 0.78&0.54 / 1.15\\
Social-bigat-20~\cite{kosaraju2019social} &0.69 / 1.29&0.49 / 1.01&0.55 / 1.32&0.30 / 0.62&0.36 / 0.75&0.48 / 1.00\\
STGAT-20~\cite{huang2019stgat}&0.65 / 1.12& 0.35 / 0.66 &0.52 / 1.10&0.34 / 0.69&0.29 / 0.60&0.43 / 0.83\\
S-STGCNN-20~\cite{mohamed2020social} & 0.64 / 1.11& 0.49 / 0.85 & 0.44 / 0.79 & 0.34 / \textbf{0.53} &0.30 / 0.48 & 0.44 / 0.75\\\hline
{S-STAGE (ours-multi)} &\multirow{2}{*}{\textbf{0.44 / 0.77}}&\multirow{2}{*}{\textbf{0.28 / 0.50}} & \multirow{2}{*}{\textbf{0.40 / 0.77}} &\multirow{2}{*}{\textbf{0.30} / 0.56}&\multirow{2}{*}{\textbf{0.20 / 0.37}}& \multirow{2}{*}{\textbf{0.32 / 0.59}}\\
$m_{min}$: [9,6,7,7,11]&&&&&\\
\hline
\end{tabular}
\caption{Multi-modal ADE$_{min}$ / FDE$_{min}$ metric for $T_{pred}$ = 12 timesteps are reported in meters.}
\label{tbl:results-multi}
\end{table*}

\begin{table*}[!htb]
\centering\small
\begin{tabular}{ c|c|c|c|c|c|c}

\hline
&ETH&Hotel&Univ&Zara1&Zara2&Avg\\\hline

\multicolumn{7}{c}{$\mathcal{M}_1$-metric $\uparrow$}\\\hline
S-STGCNN-20~\cite{mohamed2020social}& 1.00 / 1.79 &\textbf{0.77} / 1.44 & 0.71 / 1.35 &0.62 / 1.16&0.54 / 1.01 &0.73 / 1.35\\
S-STAGE ($m_{p_{max}}$) &1.17 / 2.56 &0.59 / 1.34&-&-&-&0.88 / 1.95\\ 
S-STAGE ($m_{min}$) &\textbf{1.63} / \textbf{3.44} &0.75 / \textbf{1.74}&\textbf{1.05} / \textbf{2.31}&\textbf{1.09} / \textbf{2.45}&\textbf{1.17} / \textbf{2.47}&\textbf{1.14} / \textbf{2.48}\\ \hline\hline
\multicolumn{7}{c}{$\mathcal{M}_2$-metric $\downarrow$}\\\hline
S-STAGE ($m_{p_{max}}$) &0.07 / 0.16& 0.15 / 0.33&-&-&-&0.11 / 0.24\\ 
S-STAGE ($m_{min}$) &\textbf{0.06 / 0.14}&\textbf{0.03 / 0.06}&\textbf{0.17 / 0.34}&\textbf{0.08 / 0.16}&\textbf{0.13 / 0.26} & \textbf{0.09 / 0.19}\\

\hline

\end{tabular}
\caption{$\mathcal{M}_1$ and $\mathcal{M}_2$ associated with ADE / FDE for $T_{pred}$ = 12 timesteps are reported in meters.}
\vspace{-0.5cm}
\label{tbl:results-new}
\end{table*}

We choose $\hat{e}$ (best mode error) either as $e_\mu$ (error contributed by the mean of their output distribution) when the probabilities are not available~\cite{mohamed2020social} or as $e_{p_{max}}$ (error contributed by the $p_{max}$ maximum probability mode) when the probabilities are associated with each mode like our Social-STAGE. 

\section{Experimental Results}
We evaluate 5 different scenarios on publicly available ETH~\cite{pellegrini2009you}-UCY~\cite{lerner2007crowds} datasets that have been widely used in pedestrian trajectory forecast. We use the same splits of train, validation, and test sets as in \cite{gupta2018social,mohamed2020social} for a fair comparison.

\subsection{Quantitative results}
\begin{figure}[t]
    \centering
    \includegraphics[width=1\linewidth]{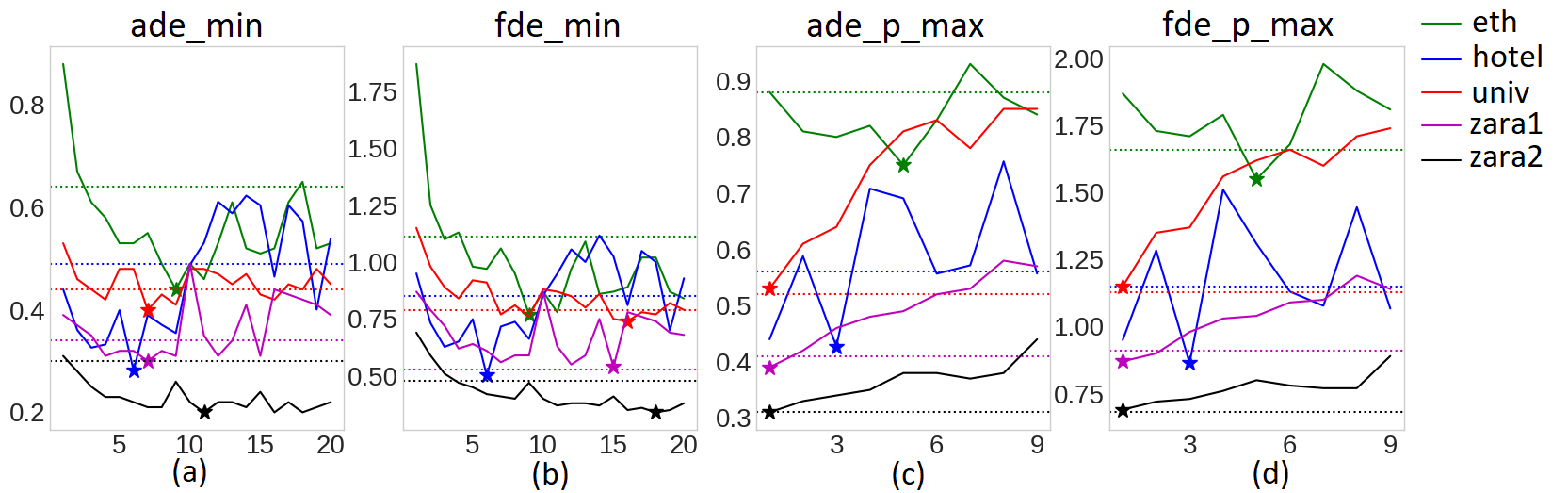}
    \caption{Modes vs Metrics: We plot number of modes on x axis and different errors (in meters) on y-axis for all 5 different sets of ETH-UCY. The dotted line shows corresponding best state-of-the-art baseline in that set and metric. Star denotes to best number of modes to represent that dataset.}
    \label{fig:modes_vs_metrics}
\end{figure}

In Table~\ref{tbl:results-single},~\ref{tbl:results-multi},~\ref{tbl:results-new}, we show different types of quantitative evaluations of our framework on ETH-UCY datasets by comparing to other state-of-the-art baselines. Errors of all baselines and our Social-STAGE (S-STAGE for simplicity) are reported in meters, with 8 time steps of observations and 12 time steps of predictions in future. In Table~\ref{tbl:results-single}, we show single-modal comparisons. Our S-STAGE (single-modal) with graph convolution module and multi-attention outperforms similar graph-based methods such as S-STGCNN~\cite{mohamed2020social} and STGAT~\cite{huang2019stgat}. Note that single modal comparisons are not reported in the S-STGCNN paper. Therefore, our evaluation is based on their  publicly available trained models and code. We use the mean from the Gaussian prediction for S-STGCNN single-modal evaluation baseline. We additionally report S-STAGE (ours $p_{max}$) that is using the best mode of trajectory (with the highest probability) for evaluation. This model performs better overall as compared to all other baselines. It highlights the efficacy of our ranking capability. We also report the best number of modes that are empirically calculated as shown in Figure~\ref{fig:modes_vs_metrics}. Here, we plot the metrics on Y-axis and modes on X-axis with corresponding best baseline in dashed line. We observe that $p_{max}$ error does not decrease with more number of modes in Figure~\ref{fig:modes_vs_metrics} (c-d).
The best performance is shown from $5$ modes for ETH and $3$ modes for Hotel. However, in case of Univ, Zara1, Zara2, the maximum probability error $m_{p_{max}}=1$ is the same as those of S-STAGE (ours-single), which seems consistent with the motion complexity of these sets. 

For multi-modal comparison, the best performing modes for ADE is different with those of FDE as shown in Figure~\ref{fig:modes_vs_metrics} (a-b). We choose best number of modes ($m_{min}$) based on minimum \textit{ADE}$_{min}$ for reporting both \textit{ADE}$_{min}$ and \textit{FDE}$_{min}$. The minimum error metrics are shown in Table~\ref{tbl:results-multi} for multi-modal prediction. Interestingly, we observe that S-STAGE performs the best even with less number of modes in all sets compared to the baseline models, which demonstrate the robustness of the proposed framework.

We also report our new metrics $\mathcal{M}_1$ and $\mathcal{M}_2$ in Table~\ref{tbl:results-new} to evaluate diversity of different modes ($\mathcal{M}_1$) and with their confidence ($\mathcal{M}_2$). For $\mathcal{M}_1$, we compare with S-STGCNN-20 using $\hat{e}=e_\mu$ as their best mode error. 
We observe in $\mathcal{M}_1$ that the model trained for the maximum probability mode, $m_{p_{max}}$ (with 5 modes for ETH), achieves better diversity than 20 modes sampled from S-STGCNN. When we compare our $m_{min}$ ([9,6,7,7,11]) model that is trained for multi-modal evaluation for this metric, it shows the way higher diversity among the predicted trajectories. For evaluation on $\mathcal{M}_2$, note that we compare our $m_{p_{max}}$ and $m_{min}$ modes baselines since the compared state-of-the-art methods do not provide probabilities associated with prediction samples. We observe that both baselines perform similar in the ETH set. But for the Hotel set, we find more modes are giving less error. 
Overall, we find that the $m_{min}$ baseline predicts high confident probabilities. 

\subsection{Qualitative results}
\begin{figure}[t]
    \centering
    \includegraphics[width=1\linewidth]{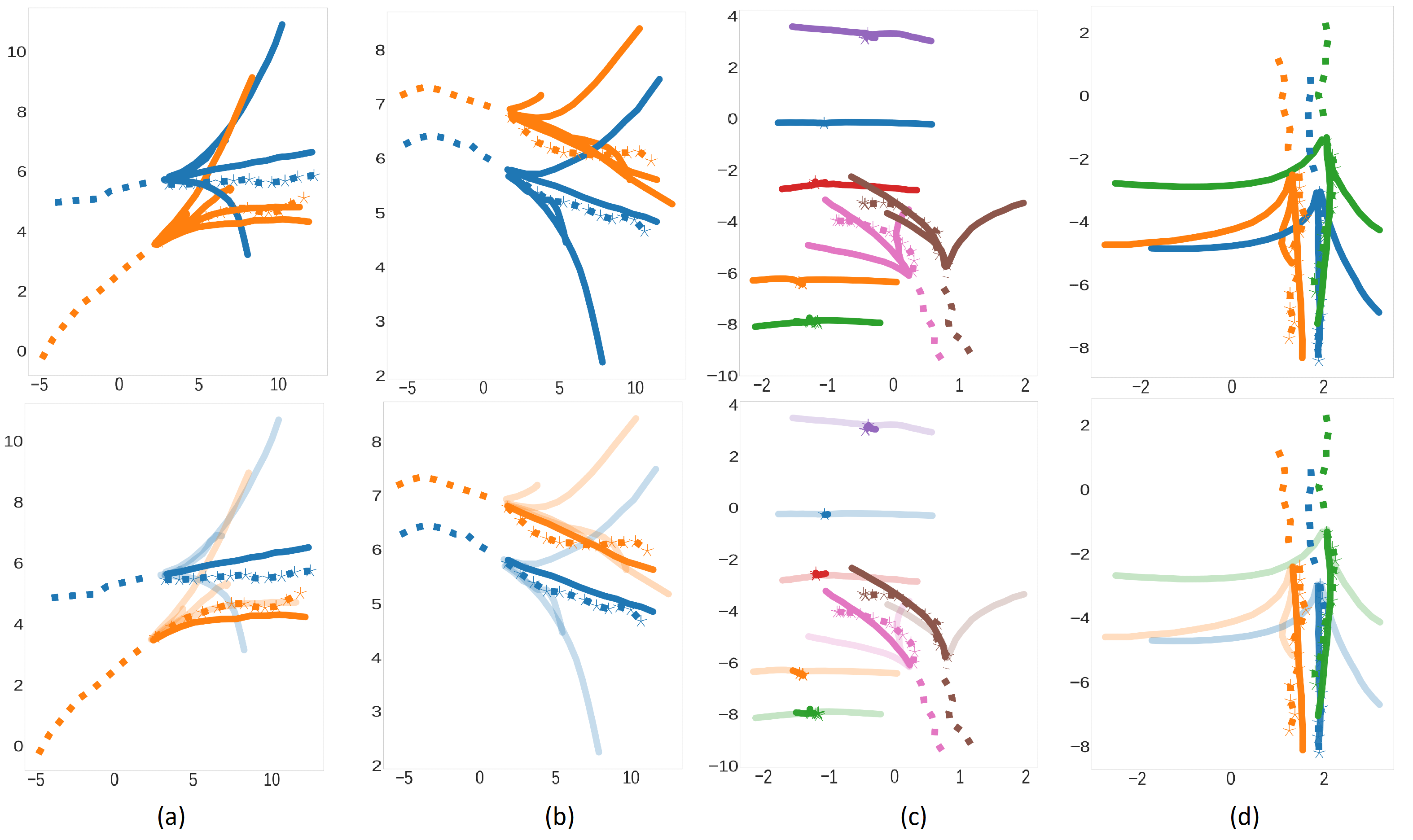}
    \caption{Social-STAGE qualitative results. Different agents are shown in different colors. First row all modes with equal probability. Second row with probability as opacity. Columns (a,b) from ETH test-set, and columns (c,d) from hotel test-set.}
    \label{fig:qualitative_results}
    \vspace{-0.5cm}
\end{figure}
We show qualitative results of our model for both multi-modal predictions and ranking in Figure~\ref{fig:qualitative_results}. Different pedestrians are visualized using different colors. Top row shows multi-modal prediction scenarios (a,b) from ETH set (with number of modes $M=5$) and (c,d) from hotel set (with number of modes $M=3$). In scenario (a,b), when two pedestrians walk closely, the predictions are made with max probability close to ground truth based on the motion cues and interaction cues. In scenario (c), there are many pedestrians standing and not moving. Based on their motion cues, the predicted max probability mode generates 'standing' in the future. Also, other modes predict that they can move in two opposite directions, which demonstrates the diversity of our proposed model. In the same scenario (c), there are two pedestrians walking beside each other. Based on motion cues, the network predicted that turning left trajectory as a dominant mode. In scenario (d), three pedestrians are walking very closely along side of each other. Our model predicts going straight as a more dominant future, but at the same time, we also predict other possibilities of turning left or right, which is common and plausible in real scenarios.

\section{Implementation and Model Details}
\label{sec:implementation_model}
We provide details of our model implementation below with the following settings: an example batch size of 1; number of agents $K$=4; number of output modes $M$=2; number of input time steps $T_{in}$=8; and number of output time steps $T_{out}$=12. We trained for 100 epochs with $M$ ranging from 0 to 20. We reported best epoch and best modes $M$ results. We used a learning rate of 0.0001 and trained our model on Quadro RTX 6000 GPU. We used the PyTorch framework for the implementation.

\begin{table}[h!]
\centering\tiny
\begin{tabular}{ c|l|c|c}
 &Layer & Kernal shape & Output shape\\
\hline\hline
&  \multicolumn{3}{|c}{Graph Convolutions}\\\hline
0&l1.BatchNorm2d & [2] & [1, 2, 8, 4]\\
1&l1.PReLU & [1] & [1, 2, 8, 4]\\
2&l1.Conv2d &[2, 2, 3, 1]  & [1, 2, 8, 4]\\
3&l1.BatchNorm2d & [2] &  [1, 2, 8, 4]\\
4&l1.Dropout & -  & [1, 2, 8, 4]\\
5&l2.Conv2d &[2, 2, 1, 1] & [1, 2, 8, 4]\\
6&l2.PReLU & [1] & [1, 2, 8, 4]\\\hline
7&\multicolumn{3}{|c}{graph conv update using adjacency matrix,}\\ 
&\multicolumn{3}{|c}{follow Equation~[2] from manuscript}\\\hline
&  \multicolumn{3}{|c}{Attention Module}\\\hline
8&attn.BatchNorm2d& [2] &  [1, 2, 8, 4] \\  
9&attn.PReLU&   [1]  & [1, 2, 8, 4] \\
10&attn.Conv2d & [2, 2, 3, 1] & [1, 2, 8, 4] \\
11 &attn.BatchNorm2d & [2] &  [1, 2, 8, 4] \\  
12 & attn.Softmax &- &  [1, 2, 8, 4] \\\hline
13&  \multicolumn{3}{|c}{using attn weights from step 12, follow} \\
&\multicolumn{3}{|c}{Equation~[3] from manuscript (multi-attention operation)}\\\hline
& \multicolumn{3}{|c}{Decoder}\\\hline
14(a)&d.Conv2d&[8, 24, 3, 3]&[1, 24, 2, 4]\\
15(a)&d.PReLU&[1] &[1, 24, 2, 4]\\
16(a)&traj.Conv2d & [24, 24, 3, 3] & [1, 24, 2, 4]\\
16(b)&prob.Conv2d& [16, 2, 3, 3] & [1, 2, 1, 4]\\
17(a)&traj.reshape & - & [1, 2, 12, 2, 4]\\\hline
\hline
\end{tabular}
 \caption{S-STAGE model summary}
  \label{tbl:fol_model}
\end{table}

$D_{in}$ is the input dimension of the trajectory at each time step, since the ETH-UCY dataset consists of 2D motion of pedestrians $D_{in}=2$. $D_{out}$ is the output dimension of the trajectory, if the prediction is Gaussian distribution for each mode (gaussian mixture) $D_{out}$ should be 5 (containing variance and correlation outputs), if the prediction is a direct trajectory regression $D_{out}$ should be 2. We present $D_{out}=2$ in this work, as we observed $D_{out}=5$ with negative log-likelihood for multi-modal setup did-not perform better than $D_{out}=2$ with rmse as a common issue of mode collapse while training mixture distributions [19,20,21,8,9,4].

\section{Conclusion}
\label{sec:conclusion}
We presented a trajectory prediction framework that aims to generate plausible future trajectories with a diversity, considering spatio-temporal interactions of agents. Given the agents’ motion history, we explored the spatial influences of individual agents and their temporal changes. We then found the relative importance of interactions using multi-attention, which can be simultaneously captured along the spatial and temporal dimension. The resulting features were used to generate multiple future trajectories with corresponding probabilities to rank each mode. To this end, we evaluated our approach using the public benchmark datasets and showed significant improvement for single- and multi-modal prediction on the standard ADE and FDE metrics as well as $\mathcal{M}_1$ and $\mathcal{M}_2$ that were newly introduced in this work to evaluate the diversity with associated probability of multi-modal prediction. 






\bibliographystyle{IEEEtran}
\bibliography{references}  
\end{document}